\documentclass[a4paper]{svproc}

\usepackage{url}
\usepackage{wrapfig}
\usepackage{graphicx}
\usepackage{algorithm}
\usepackage{amsmath}
\usepackage{algpseudocode}
\usepackage{amssymb}
\usepackage{amsfonts}
\usepackage[T1]{fontenc}
\usepackage[english]{babel}

\usepackage[T1]{fontenc}
\usepackage[english]{babel}
\usepackage{amssymb}
\usepackage{xcolor}
\usepackage[numbers,sectionbib]{natbib}
\usepackage{subfig}
\usepackage{paralist}
\usepackage{enumitem}
\usepackage{siunitx}

\newcounter{daggerfootnote}

\makeatletter
\renewcommand{\Function}[2]{%
  \csname ALG@cmd@\ALG@L @Function\endcsname{#1}{#2}%
  \def\jayden@currentfunction{#1}%
}
\newcommand{\funclabel}[1]{%
  \@bsphack
  \protected@write\@auxout{}{%
    \string\newlabel{#1}{{\jayden@currentfunction}{\thepage}}%
  }%
  \@esphack
}
\makeatother
\begin{document}

\mainmatter              
\title{Active Rendezvous for Multi-Robot Pose Graph Optimization using Sensing over Wi-Fi}
\author{Weiying Wang\inst{1}{$\dagger$}  \and Ninad Jadhav\inst{1}{$\dagger$} \and Paul Vohs\inst{1} \and Nathan Hughes\inst{2} \and Mark Mazumder\inst{2} \and Stephanie Gil\inst{1}{$\dagger$} }

\titlerunning{Active Rendezvous for PGO using Wi-Fi}  
%
%
\authorrunning{Wang et al.} 
%
\tocauthor{Weiying Wang, Ninad Jadhav, Paul Vohs, Nathan Hughes, Mark Mazumder, Stephanie Gil}
\institute{Arizona State University, Tempe AZ 8500  4, USA,\\
\and
MIT Lincoln Laboratory, Lexington, MA 02421, USA}
   \begin{center}
   \begin{scriptsize}
    Accepted for publication at ISRR 2019, please cite as follows: \\
    W. Wang, N. Jadhav, P. Vohs, N. Hughes, M. Mazumder, S. Gil \\
    Active Rendezvous for Multi-Robot Pose Graph Optimization using Sensing over Wi-Fi \\ The International Symposium on Robotics Research 2019
   \end{scriptsize}
   \end{center}
\begin{minipage}{\textwidth}
   \maketitle
\end{minipage}
\vspace{-0.8cm}
\begin{abstract}
We present a novel framework for collaboration amongst a team of robots performing \emph{Pose Graph Optimization} (PGO) that addresses two important challenges for multi-robot SLAM: i) that of enabling information exchange ``on-demand'' via \emph{Active Rendezvous} without using a map or the robot's location, and ii) that of rejecting outlying measurements. Our key insight is to exploit relative position data present in the communication channel between robots to improve groundtruth accuracy of PGO. We develop an algorithmic and experimental framework for integrating \emph{Channel State Information }(CSI) with multi-robot PGO; it is distributed, and applicable in low-lighting or featureless environments where traditional sensors often fail. We present extensive experimental results on actual robots and observe that using \emph{Active Rendezvous} results in a 64\% reduction in ground truth pose error
and that using CSI observations to aid outlier rejection reduces ground truth pose error by 32\%.  These results show the potential of integrating communication as a novel sensor for SLAM.


\vspace{-0.3cm}
\vspace{-0.6cm}
\end{abstract}
\vspace{-0.9cm}
\section{Introduction}
\vspace{-0.3cm}
\begin{wrapfigure}[13]{r}{0.45\textwidth}
  \begin{center}
  \vspace{-38pt}
    \includegraphics[width=0.42\textwidth]{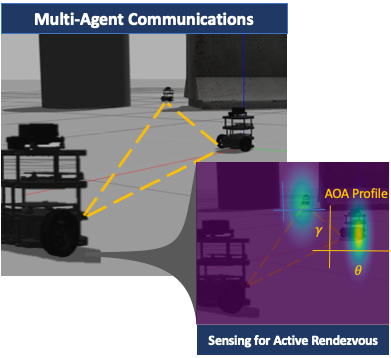}
  \end{center}
  \vspace{-20pt}
  \caption{\footnotesize{\emph{Sensor information from wireless signals used for multi-robot PGO.}}}
  
\end{wrapfigure}
\let\thefootnote\relax\footnotetext{{$\dagger$}Moved to Harvard University as of 2020\\ \{weiyingwang, njadhav\}@g.harvard.edu, sgil@seas.harvard.edu}
Multi-robot mapping allows for quick exploration of unknown environments and resiliency to individual robot failures.  There has been a significant effort to establish the necessary coordination algorithms for efficient and accurate mapping using information fused from teams of robots~\cite{multiRobotSurvey,distributedMappingFox,coordinationMapping,RoyRendezvous}.
Data exchange is a critical need for successful execution of a distributed mapping task, both for map fusion and for reducing errors in pose estimates that accumulate over time due to noise. A common assumption in these works is that robots can share information reliably and that relative position information is sufficiently accurate to allow for integration of data~\cite{distributedMappingFox, multiRobotSurvey}; assumptions that are challenging to uphold in practice. For information exchange, random rendezvous opportunities are often exploited, or it is assumed that robots can navigate to a common location in the map to rendezvous~\cite{RoyRendezvous,particleFilter,multiRobotSurvey}.  However, the question of how to enforce a communication rendezvous in a \emph{distributed} fashion without requiring robots to have an agreed upon shared map remains a challenge. Additionally, outlying relative pose measurements between robots can result from many causes including erroneous loop closure, data mismatch, and sensor noise amongst others. These outlier measurements can be severely detrimental to accuracy~\cite{choudhary2017distributed}. Detecting such outliers is challenging and has been identified as an open problem for the next era of multi-robot simultaneous localization and mapping (SLAM), penned the ``Robust Perception Age''~\cite{lucaSurvey}.\\  
\indent Ideally, it would be possible to address both of these challenges using local information that already exists in the system but is not currently exploited for PGO.  By building upon a key insight, that existing communication links between robots contain relative position information that can be fruitfully exploited in PGO, we address these challenges from a new perspective.\\ 
\indent We present an algorithmic framework where each robot monitors its trajectory error and its wireless channels to neighboring robots in order to proactively request a rendezvous whenever this error exceeds a prescribed threshold. By enabling information exchange, robots are able to improve the groundtruth accuracy of their estimated pose obtained with PGO. We refer to this as \emph{Active Rendezvous}. We accomplish this by building upon our prior work that uses Channel State Information (CSI) observations collected from communication links between robots to derive the \emph{angle-of-arrival} (AOA) for received signals without specialized hardware~\cite{IJRR,ubicarse}. Whereas~\cite{IJRR} uses a global optimization approach to improve communication across a network of robots, the current work differs in several key aspects necessary for integration with PGO: it 1) is entirely \emph{distributed}, allowing robots to follow gradient directions of communication quality with a desired neighboring robot 2) identifies a neighboring robot for rendezvous that would best reduce robot's trajectory error, and 3) modifies the PGO formulation to use inter-robot position data measured from communication channels to reduce the influence of outliers. To address outliers, we devise a mathematical framework for using relative pose information from the inter-robot communication signals to scale the information matrix of each relative measurement. When used by PGO, this scaling automatically reduces the influence of outliers.  Since PGO is often performed by robots operating in indoor environments that have structure, or \emph{multipath} generators, any method for using information from communication signals to reject outliers must i) handle multipath and ii) characterize the accuracy with which it can validate reported relative positions in these environments. Our framework specifically handles the case of multipath, leading to good performance across different testbeds with and without obstacles (multipath generators), and we provide an experimental characterization of error in AOA needed for outlier detection (Figure~\ref{fig:csi_rejection} (b) and (c)).  In this way our method \emph{informs} outlier rejection effectively by using CSI as an independent source of validation for data shared between robots; Thereby improving the resulting accuracy of PGO. \\
\indent By integrating AOA information from communication channels with PGO, we address two important problems: 1) realizing data exchange on demand, or \emph{Active Rendezvous}, between robots to keep trajectory error below a desired threshold, and 2) mitigating the effect of outliers on PGO to improve the solution accuracy. Our method is environment independent, distributed, and compatible with any exploration method where information exchange is critical. Our experiments show that incorporating AOA information significantly reduces trajectory error, leading to a 64\% reduction of pose error with respect to ground truth. Notably, our methods are applicable to environments where conventional sensing modalities may fail. As such, this paper begins to lay a foundation for using communication-as-a-sensor for multi-robot mapping. \\
\vspace{-0.85cm}
\section{Related Work}
\vspace{-0.35cm}
While the body of related work for distributed mapping and Pose Graph Optimization (PGO) is too large for a comprehensive overview, we survey the most relevant works to current paper objectives here.
PGO is one of the most common modern techniques for SLAM, and supports many crucial tasks in robotics~\cite{lucaSurvey}.Techniques for distributed PGO include reducing the amount of poses communicated~\cite{choudhary2017distributed}, the use of graph sparsification~\cite{kretzschmar2012information,paull2015communication}, and efforts to reduce bandwidth requirements~\cite{cunningham2012fully,lazaro2013multi}. However, these techniques still suffer from outlying observations and the inability to coordinate information exchange on demand. This paper builds off of distributed PGO techniques as in~\cite{choudhary2017distributed} but incorporates information from communication channels to alleviate these two challenges. Another approach to these challenges, Active SLAM, attempts to incorporate estimation uncertainty into the control of robots, such as estimating information-gain from exploration versus uncertainty reduction via repeated visitations~\cite{vallve2015active,valencia2012active} using pose graphs. Our approach is compatible with different exploration methods of the environment, such as Active SLAM. \\
\indent Localization and pose estimation using various Radio Frequency (RF) measurements comprises a large body of work~\cite{khan2015linear,lorincz2005motetrack,wei2014noisy,todescato2015distributed,piovan2013frame}. Unfortunately, these methods often require specialized hardware that is not normally found on robots, or are marred by the phenomenon of \emph{multipath} which is often present in indoor environments of interest in mapping tasks.  Most related to our approach are synthetic aperture radar (SAR) techniques for indoor positioning that can handle multipath and alleviate the need for bulky multi-antenna arrays~\cite{ubicarse,lteye,IJRR,PinIt}. These approaches show great promise for robotic applications due to their compatibility with Wi-Fi, which is present on many robot platforms. \\
\indent The current paper builds atop this body of related work, in particular~\cite{IJRR}, and extends it in several critical ways necessary for multi-robot PGO. In contrast to~\cite{IJRR}, this work is distributed and derives decentralized methods for reducing pose estimation error based on \emph{trajectory error} and real-time \emph{Effective Signal to Noise Ration} (ESNR) values.
In the context of other outlined mapping and localization approaches, this paper highlights an algorithmic and experimental framework to intelligently incorporate communications-channel information into collaborative mapping techniques. 
\vspace{-0.45cm}
\section{Problem Statement}
\vspace{-0.3cm}
We consider problems where $n$ robots explore an unknown environment and must maintain accurate estimates of their poses. To prevent divergence of the estimation process, these robots must periodically exchange information via rendezvous. Our approach does not make assumptions about \emph{how} robots explore the environment and is independent of the exploration strategy so long as regular information exchange is required during exploration. Here, robot $i$'s pose is given by $x_i(t) := \left( R_i(t), p_i(t) \right)$ where $p_i(t)\in \mathbb{R}^3$ is the position of robot $i$ at time $t$ and $R_i(t)\in SO(3)$ is the orientation of robot $i$ at time $t$. $SO(3)$ here is the special orthogonal group, defined as $\{R\in\mathbb{R}^{3\times 3}:R^{\intercal} R=I_3, det(R)= +1\}$. We assume that information exchange, or communication rendezvous, between two robots $i$ and $j$ can occur whenever a communication link exists between robots that can support the required data rate $q_i\geq 0$. This is the required data rate for exchanging information to reduce the estimation error of the robot. In practice this value is dependent on the nature of the data exchanged, but most commonly includes position estimates and a history of observations of the environment. It is important to note that the minimum communication rate required to ``sense'' or broadcast ping packets is often possible at much larger distances than the communication rate required for a rendezvous. We define the set of robots that robot $i$ has minimal connectivity to as $\mathcal{N}_{C_i}(t)$.  We leverage previous work on attaining directional signal profiles, or \emph{angle-of-arrival} information for each communicating pair of robots~\cite{IJRR} inside $\mathcal{N}_{C_i}(t)$. A primer on how to attain the AOA profile $F_{ij}$ for each pair of communicating robots is included in Sec.~\ref{sec:primer}. The goal of this paper is to use these AOA profiles between communicating robots to improve the accuracy of the PGO process.  

\vspace{-0.45cm}
\subsection{Pose Graph Optimization}
\label{sec:PGO}
\vspace{-0.25cm}
We assume that each robot \emph{i} acquires relative pose measurements $\bar{z}_{j}^i(t)$, i.e the pose of robot $j$ with respect to robot $i$'s reference frame at time $t$.
We follow the observation model from~\cite{choudhary2017distributed}, dropping $t$ for clarity:
\begin{align}
\label{eq:observation_model}
\bar{z}_j^i := (\bar{R}_j^i,\bar{p}_j^i), \text{ where } \bar{R}_j^i=(R_i)^ {\intercal} R_j R_\epsilon \text{ and } \bar{p}_j^i=(R_i)^{\intercal}(p_j-p_i)+p_\epsilon
\end{align}
Here $(R_{\epsilon}, p_{\epsilon})$ denotes the noise of the observation and, roughly speaking, is assumed to be drawn from some zero-mean Gaussian distribution with Fisher information matrix $\Omega_{\bar{z}_j^i}$. 
We define $\mathcal{E}_{\bar{z}}(t)$ as the set of all relative pose observations for all robots that occur up to time $t$, and $\mathcal{E}_{\bar{z}}^i(t)$ as the subset of $\mathcal{E}_{\bar{z}}(t)$ relative to robot $i$.
We assume the robots are capable of an estimation process to find an optimized set of poses $x^*$ following the maximum likelihood formulation: 
\setlength{\belowdisplayskip}{3pt}
\setlength{\abovedisplayskip}{3pt}
\begin{equation}
\label{eq:PGO}
x^*=\arg\max_x \prod_{\bar{z}\in\mathcal{E}_{\bar{z}}(t)} L(\bar{z}|x)
\end{equation}

\noindent Our development is largely independent of the method to obtain $x^*$ (we refer to~\cite{choudhary2017distributed, graphSLAMtutorial} for examples) and thus we keep the concepts general where possible. We assume the method to obtain $x^*$  allows for any robot $i$ to compute the optimization cost for all measurements relative to itself at any time, denoted as $Err_i(x_i(t), \mathcal{E}^i_{\bar{z}}(t))$, which we refer to as \emph{``trajectory error''}. One such example from~\cite{choudhary2017distributed}, which is used in our experimental evaluation, is:
\setlength{\belowdisplayskip}{3pt}
\setlength{\abovedisplayskip}{3pt}
\vspace{-0.1cm}
\begin{equation}
    \label{eq:internal_pose_error}
    Err_i(x_i(t), \mathcal{E}_{\bar{z}}^i(t))) := \sum_{\bar{z}\in\mathcal{E}_{\bar{z}}^i(t)} \omega_p^2 ||p_j - p_i- R_i\bar{p}_j^i||^2 + \frac{\omega_R^2}{2}||R_j - R_i\bar{R}_j^i||_F^2
\end{equation}
where $||\cdot||^2_F$ is the Frobenius norm and $\omega_p^2$ and $\frac{\omega_R^2}{2}$ are concentration parameters for the distribution of $R_{\epsilon}, p_{\epsilon}$. 
 
Our goal is to ensure that once the trajectory error passes a user defined threshold $\delta>0$, a subset of the robots can engage in \emph{Active Rendezvous} to gather and exchange information to reduce overall estimation errors. We stress that the concepts of this paper are extensible beyond Equation~\eqref{eq:internal_pose_error} to any error metric $Err_i$ that can be computed directly from the estimation process.
Therefore, our objectives are two-fold: for every robot $i \in \left\{1, \ldots, n\right\}$ with the local observations $\mathcal{E}_{\bar{z}}^i(t)$ and observations over the wireless channel of $F_{ij}$ for every robot $j$ in $\mathcal{N}_{C_i}(t)$, we aim to address:
\vspace{-0.1cm}
\begin{problem}[Active Rendezvous]
	Develop an algorithm for achieving \emph{Active Rendezvous} between robots that is distributed and independent of the acquired map.
\end{problem}
\vspace{-0.4cm}
\begin{problem}[Outlier Rejection]
	Develop a framework that uses observations over the wireless channels, $F_{ij}$, to mitigate outlying observations in $\mathcal{E}_{\bar{z}}(t)$.
\end{problem}

\vspace{-0.7cm}
\section{Primer: Angle-of-Arrival in Multi-Robot Systems}
\label{sec:primer}
\vspace{-0.30cm}
\setlength{\belowcaptionskip}{-10pt}

\begin{wrapfigure}[13]{r}{0.5\textwidth}
  \begin{center}
 \vspace{-37pt}
    \includegraphics[width=0.48\textwidth]{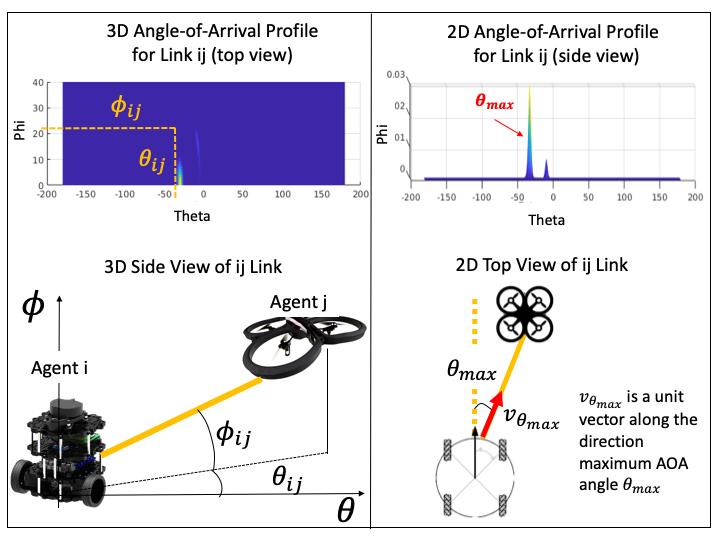}
  \end{center}
 \vspace{-13pt}
  \caption{\footnotesize{\emph{AOA information from a signal profile using Channel State Information (CSI).}}}
  \label{fig:aoa}
\end{wrapfigure}
\setlength{\belowcaptionskip}{3pt}

As robots communicate, messages are transmitted over wireless channels $h$ which are complex numbers exposed by some commodity Wi-Fi cards and measurable from any wireless device. These quantities characterize the power and phase of a received signal, and are affected by the path the signal traverses (attenuation) and the environment (scattering, multi-path), which can reveal information about the direction of the signal's source.  Attaining (AOA) information requires hardware that is normally too bulky for small robots, but previous work demonstrates that the combination of robot motion and wireless channel measurements can instead be used to obtain this information. By having robot $i$ take several snapshots of the wireless channel $h_{ij}$ 
to robot $j$ as it moves over a piece-wise linear~\cite{IJRR} or circular trajectory~\cite{AURO2017}, a \emph{signal profile} can be computed as:
\begin{align}
\label{eq:fij}
&F_{ij}(\phi,\theta)=\frac{1}{\left|Eig_n(\hat{h}_{ij}\hat{h}_{ij}^+)\exp^{\sqrt{-1}\Phi_{ij}(\phi,\theta)}\right|^2}
\end{align}
\noindent An actual measured profile $F_{ij}$ is shown in Fig.~\ref{fig:aoa} for $\theta \in (-180,180)$ and $\phi\in(0,40)$), and the unit vector $v_{\theta_\text{max}}(F_{ij})$ along the direction $\theta_\text{max}$ is depicted in the schematic on the right hand side of Fig.~\ref{fig:aoa}.  This captures the direction of arrival of the signal in 3D space between two communicating robots $i$ and $j$ where $\phi$ is the polar direction of the arriving signal (out of the plane) and $\theta$ is the azimuthal direction of the arriving signal (in the plane). Here the raw channel ratio between two antennas is used to define $\hat{h}_{ij}=h_{1ij}/h_{2ij}$, a vector of the ratio of wireless channel snapshots between the two receiving antennas mounted on robot $i$ (Fig.~\ref{fig:hardware_setup} shows a typical hardware setup). The function $\Phi_{ij}(\phi,\theta)$ is defined as $\Phi_{ij}(\phi,\theta)=\frac{2\pi r}{\lambda}\cos(\phi-B_j)\sin(\theta-\Gamma_j)$, $\lambda$ is the wavelength of the signal, $r$ is the distance between the two antennas, $B_j,\Gamma_j$ are the transmitters orientation, $Eig_n(\cdot)$ are noise eigenvectors, $(\cdot)^+$ is the conjugate transpose, and $k$ is the number of signal eigenvectors, equal to the number of paths (in practice $k$ is related to the number of peaks in the profile~\cite{AURO2017}). 
In the current paper robot $i$ performs a quarter turn (circular motion) while obtaining wireless channel snapshots once during each iteration of Algorithm~\ref{alg:rendezvous} to obtain $F_{ij}$.
Of particular importance to our approach is prior work which compares peak locations in the $F_{ij}$ profile to derive a likelihood that a reported relative position between the robots $i$ and $j$ is correct~\cite{AURO2017}.  Let $\mathcal{W}_{ij}(\phi_{ij},\theta_{ij})$ be the likelihood that a transmitting robot $i$ is indeed along its reported direction $(\phi_{ij},\theta_{ij})$ with respect to a receiving robot $j$. The likelihood is derived as (cf. $\beta$ in~\cite{AURO2017}): 
\begin{equation}
\label{eq:w}
    \mathcal{W}_{ij}(\phi_{ij},\theta_{ij})=g(\phi_{ij}-\phi_{F_{ij}};0,\sigma_\phi^2)\times g(\theta_{ij}-\theta_{F_{ij}};0,\sigma_\theta^2)
\end{equation}

\noindent where $\phi_{F_{ij}}$ and $\theta_{F_{ij}}$ denote the closest maximum in $F_{ij}$ to $(\phi_{F_{ij}},\theta_{F_{ij}})$ and\\ $g(x;\mu,\sigma^2)=\min(1,\sqrt{2\pi}f(x;\mu,\sigma))$ is a normalized Gaussian PDF with mean $\mu$ and variance $\sigma^2$.  In this paper we will provide an experimental study of $\mathcal{W}_{ij}$ and we will develop the necessary framework for integrating this likelihood into pose-graph optimization so that AOA can be used as an independent observation to detect outliers in relative position measurements between robots.

\vspace{-0.4cm}
\section{Active Rendezvous}
\label{sec:rendezvous}
\vspace{-0.4cm}

In this section we describe a framework for achieving \emph{Active Rendezvous} between robots to keep their trajectory errors below a desired threshold. This threshold is arbitrary, specified by the user, and often related to how much trajectory error is tolerable for the task at hand.
As robots move through the environment they accumulate odometer and sensor errors~\cite{graphSLAMtutorial}. By exchanging information, such as a history of observations and relative pose estimates, these errors can be reduced. We thus introduce decentralized methods for: 1) initiating a request for \emph{Active Rendezvous} among robots when their estimated trajectory errors grow beyond an acceptable threshold, 2) selecting the subset pair of robots to be activated to complete an \emph{Active Rendezvous} based on ESNR and trajectory error values, and 3) the set of relative position commands, based on wireless signal profiles, to be executed in a distributed manner to achieve rendezvous.  Before detailing our methods, we define different robot behaviors below:
 \begin{enumerate}[noitemsep, nolistsep,topsep=0pt]
\item $move\_to\_relative\_waypoint$: robot moves toward a given waypoint. 
\item $exploration$: arbitrary path finding policy to explore the environment. 
\item $adaptive\_walk$: series of $move\_to\_relative\_waypoint$ using signal profiles.

\end{enumerate} 
\setlength{\textfloatsep}{0pt}
\begin{algorithm}[h]
\caption{ERROR MONITOR}
\label{alg:monitor}
\begin{algorithmic}[1]
\Require{ Estimated pose error threshold $\delta$, robot index $i$, desired data exchange rate $q_{i}$, minimum number of observations $\kappa$, connected neighbors of $i$, $\mathcal{N}_{C_i}$, $\alpha_i$ for each robot(can be all same or arbitrary value based on priority of robot by demand) }
\State $t \leftarrow 1, x_{i} \leftarrow \emptyset, \mathcal{E}_{\bar{z}}^i \leftarrow \emptyset $ \Comment{Initialization}
\While {$explore$}
    \State $[x_{i}(t), \mathcal{E}_{\bar{z}}^{i}(t)] \leftarrow \emph{$exploration$}$ 
    \State $Err_i(x_i(t),\mathcal{E}_{\bar{z}}^i(t))$ $\leftarrow$ Equation~\eqref{eq:internal_pose_error} \label{line:err}
    \If{$Err_i(x_i(t),\mathcal{E}_{\bar{z}}^i(t))>\delta$}
        \State $min\_service\_discrepancy \leftarrow \infty$
        \State robot i broadcasts \emph{active rendezvous} requests to all robots of $\mathcal{N}_{C_i}(t)$
        \For{j in $\mathcal{N}_{C_i}(t)$} \Comment{Selection of adaptive robot j}
            \If{request accepted by robot $j$}
            \State $\rho_{ij} \leftarrow ESNR_{ij}$
            \State $w_{ij} \leftarrow {max(\alpha_i \frac{q_i-\rho_{ij}}{q_i},0)}$
            \If{$w_{ij} < min\_service\_discrepancy $} 
            \State $j^* \leftarrow j$
            \State $min\_service\_discrepancy \leftarrow w_{ij}$
            \EndIf
            \EndIf
        \EndFor
        \State $x^*_{i} \leftarrow Algorithm~\ref{alg:rendezvous} (q_{i}, \kappa, j^*, x_{i})$
        \State $x_{i} \leftarrow x^*_{i}$
    \EndIf
    \State $t \leftarrow t+1$
\EndWhile
\end{algorithmic}
\end{algorithm}

\indent We assume that each robot $i$ runs Algorithm~\ref{alg:monitor} to monitor it's trajectory error (see Equation~\ref{eq:internal_pose_error}). This error function could be replaced with any generic error metric that measures the disagreement between pose estimates and measurements. Once this error passes a prescribed (arbitrary) threshold value of $\delta$ for any robot $i$ , it broadcasts minimal data packets to identify neighboring robots who can listen and reply back, to form  $\mathcal{N}_{C_i}(t)$. Each neighbor of robot $i$ will potentially receive multiple requests.  It will choose to respond \emph{true} to the requester to whom it has highest ESNR and is thus most suited to serve. In the case of ties, it will use the requesters' trajectory error as the tie breaker. Robot $i$ then uses Algorithm~\ref{alg:monitor} (line 6-17) to identify a robot $j \in \mathcal{N}_{C_i}(t)$ for a rendezvous that has the lowest service discrepancy to robot $i$, using lower trajectory error as a tie-breaker if size($\mathcal{N}_{C_i}(t)$ ) > 1. \emph{Service discrepancy}, the difference between the desired data exchange rate $q_i$ and current data exchange rate, is defined as $w_{ij}=\max(\frac{\alpha_i (q_i-\rho_{ij})}{q_i},0)$
where $\alpha_i\geq 0$ is the priority of robot $i$, and $\rho_{ij}\geq 0$ is the actual signal strength of the link between $i$ and $j$.  A service discrepancy $w_{ij}=0$ denotes that robot $j$ can transmit data to robot $i$ at its desired data rate $q_i$.
To achieve a rendezvous via Algorithm~\ref{alg:rendezvous}, robot $j$ moves towards robot $i$ which is stationary by following the gradient of the communication edge cost. This edge cost is given by $r_M(p_j,P_t,w_{ij},F_{ij})=(p'_{ij}-p_j)^TM_{ij}(p'_{ij}-p_j)$ which is a Mahalanobis distance between an robot $j$ with position $p_j$ and the ``virtual'' position, $p'_{ij}$, of robot $i$ as determined from the AOA profile $F_{ij}$~\cite{IJRR}.  Here, $p'_{ij}=p_j+w_{ij}v_{\theta_\text{max}}$ is the perceived distance between robot $i$ and $j$ as estimated from the wireless profile $F_{ij}$ and $v_{\theta_\text{max}}$ is a vector along the maximum AOA of the signal between $i$ and $j$ (see Figure~\ref{fig:aoa}), and $w_{ij}$ is the current service discrepancy of robot $j$ to robot $i$. Matrix $M_{ij}$ encodes the AOA information from the profile $F_{ij}$ as $M_{ij}=Q_{ij}\Lambda Q^T_{ij}$ where $Q_{ij}=[v_{\theta_\text{max}},v_{{\theta_\text{max}}_\perp}]$, $v_{\theta_\text{max}}$ is a unit vector along the direction of maximum AOA (see Fig.~\ref{fig:aoa} for a description of $v_{\theta_\text{max}}$), and $\Lambda=diag(\frac{1}{\sigma^2},1)$ is a diagonal matrix capturing the noise characteristics of $F_{ij}$, and $\perp$ denotes the orthogonal vector.  Minimizing this edge cost with respect to $p_j$ returns a position for robot $j$ that improves the communication rate between $i$ and $j$. The paper~\cite{IJRR} shows how to derive the value of $\sigma^2$ from the profile $F_{ij}$ however for simplicity we assume that $\sigma<1$, which amounts to the link $ij$ having more signal than noise.
Note that following the gradient of the communication edge cost does not require line-of-sight observations. During each iteration of Algorithm~\ref{alg:rendezvous}, the robot performing the rendezvous repeats this process to obtain a new position $p_j(t+1)$ of robot $j$ until a desired number $\kappa$ of relative pose observations have been generated between itself and robot $i$ for optimization. Post optimization, the robots revert back to \emph{$exploration$}; gathering odometer measurements and monitoring the trajectory error via Algorithm~\ref{alg:monitor}. 

\vspace{-0.6cm}
\begin{algorithm}
\caption{ACTIVE RENDEZVOUS}
\label{alg:rendezvous}
\begin{algorithmic}[1]
\Require{Desired data exchange rate $q_i$, minimum number of observations $\kappa$, robot $j$ selected for rendezvous, raw pose estimates $x_{i}$, some small positive values $\varepsilon>0$ }
\Ensure{Optimized pose estimates $x_i^*$, for robot $i$}
\While {$ \vert \mathcal{E}_{\bar{z}}^i(t)\vert < \kappa $} \Comment{perform \emph{adaptive\_walk}}
    \State $w_{ij}\leftarrow \max\{0,\alpha_j\frac{q_i-\rho_{ij}}{q_i}\}$ \Comment{compute service discrepancy}
    \State $F_{ij}\leftarrow \text{ Equation~\eqref{eq:fij}}$
    \State $p'_{ij}\leftarrow p_{j}(t)+w_{ij}\vec{v_{\theta_\text{max}}}(F_{ij})$ \Comment{virtual client position}
    \State $\nabla r_M=M_{ij}(p'_{ij}-p_j)$ \Comment{Compute communication gradient direction}
    \State $p_{j}(t+1)=p_{j}(t)+w_{ij}\nabla r_M$
    \State $[x_{i},\mathcal{E}_{\bar{z}}^i(t)] \leftarrow move\_to\_relative\_waypoint(p_{j}(t+1)) $
    \State $\rho_{ij} \leftarrow \text{ESNR}_{ij}$ \Comment{actual received signal strength over link $ij$}
    \If{$q_{i}\leq\rho_{ij}$}  
            \State $q_i \leftarrow q_i + (\rho_{ij} - q_i) + \varepsilon$ \Comment{Update data transfer rate until $\vert \mathcal{E}_{\bar{z}}^i(t)\vert\geq\kappa$}
         \EndIf
\EndWhile
\State $x^*_{i} \leftarrow$ Equation~\eqref{eq:PGO}
\State \Return $x^*_{i}$
\end{algorithmic}
\end{algorithm}

\section{Outlier Rejection}
\label{sec:outliers}
\begin{wrapfigure}{r}{0.35\textwidth}
  \centering
    \includegraphics[width=0.35\textwidth]{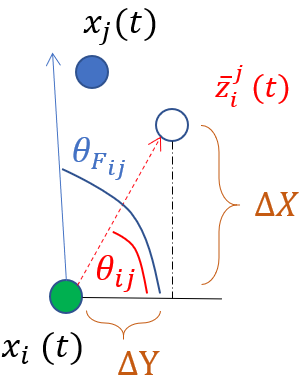}
  \caption{\footnotesize
{\emph{Angle from $F_{ij}$ versus angle from reported relative position $\theta_{ij}$ (2D).}}}
 \vspace{-23pt}
  \label{fig:Fij}
\end{wrapfigure}
Integration of relative position data from the communication channels into the PGO framework has two distinct advantages. Not only can this information be used to enable frequent data exchange opportunities as described in Section~\ref{sec:rendezvous}, but it can also be used to \emph{validate} this data thus reducing the effect of outliers in the optimization. 
This section discusses how wireless communication channels can be used to reject such outliers.
In particular, we devise a mathematical framework for using relative pose information captured in $F_{ij}$ to scale the information matrix of each relative pose observation $\bar{z}^i_j(t)$ such that outlier measurements have less influence on the final maximum likelihood estimate. Our observational model in Equation~\eqref{eq:observation_model} assumes all observations are drawn from a Gaussian distribution. An \emph{outlier} measurement however is not necessarily drawn from the same distribution model and can introduce significant bias to an optimization \cite{carlone2018convex}. Note that we make no assumptions about how outliers are generated. Ideally it would be possible to automatically mitigate the effect of outlier measurements in the optimization of Equation~\eqref{eq:internal_pose_error} by setting their information values to small or even zero values.  The AOA information captured in an $F_{ij}$ profile can be used for this purpose. Specifically, we want to verify that the observation $\bar{z}_j^i(t)$ is consistent with the observation from the AOA profile $F_{ij}$. We define $(\phi_{ij}, \theta_{ij})$ to be the relative position of robot $j$ with respect to robot $i$ which can be derived from the relative pose estimate $\bar{z}_j^i(t)$ (Fig.~\ref{fig:Fij}) as $\theta_{ij}=\arctan2(\Delta Y, \Delta X)$ and $\phi_{ij}=\arctan2(\Delta Z, \sqrt{\Delta X^2 + \Delta Y^2}\ )$. Importantly, $F_{ij}$ contains AOA information for \emph{all} incoming signal paths from $i$ to $j$ and thus can handle indoor, or \emph{multipath}, environments.  A signal path along the direction $(\phi,\theta)$ would result in a peak value of $F_{ij}(\phi,\theta)$.  Therefore we can define the closest maximum peak location in $F_{ij}$ to the reported relative position $(\phi_{ij},\theta_{ij})$ as $(\phi_{F_{ij}},\theta_{F_{ij}})$ where
\setlength{\belowdisplayskip}{3pt}
\setlength{\abovedisplayskip}{3pt}
\begin{align}
\label{eq:thetaF}
    [\phi_{F_{ij}},\theta_{F_{ij}}]=&\arg\min_{\phi,\theta}\ \alpha + \beta\\
     s.t.\ \ \ \ & \| \theta - \theta_{ij}\| \leq \alpha \nonumber, 
         \| \phi - \phi_{ij}\| \leq \beta \nonumber,   (\phi,\theta)\in \Theta_N(F_{ij}) \nonumber
\end{align}
Here $\Theta_N(F_{ij})$ is the set of angles producing the $N$ largest peaks in $F_{ij}$.  In other words, $(\phi_{F_{ij}},\theta_{F_{ij}})$ is the tuple of angles among the peaks of $F_{ij}$ corresponding to the signal direction that is closest to the reported position $(\phi_{ij},\theta_{ij})$. From Equation~\eqref{eq:w} in Section~\ref{sec:primer} we can compute the likelihood that robot $ij$ is indeed at the reported relative location $(\phi_{ij},\theta_{ij})$ according to the signal profile $F_{ij}$ as $W_{ij}\in[0,1]$.  This likelihood serves as a natural candidate for a weighting term that can be applied to the information matrix $\Omega_{\bar{z}_j^i}$ of each relative measurement between robots.  This would result in a smaller weight, nearing zero, for all relative pose measurements $(\phi_{ij},\theta_{ij})$ that are likely to be outliers as reported by the AOA signal profile with computable variance $\sigma_\theta, \sigma_\phi$. Relative pose estimates $(\phi_{ij},\theta_{ij})$ are deemed to be likely outliers if their difference to peaks in the AOA profile $F_{ij}$ is larger than a threshold value $\Delta$. Thus, for all relative pose observations, we compute a new information matrix $\Omega_{\bar{z}_j^i}'$ as:
\begin{align}
\label{eq:omega}
\Omega_{\bar{z}_j^i}' :=
\begin{cases} 
      \mathcal{W}_{ij} \cdot \Omega_{\bar{z}_j^i} & |\theta_{ij}-\theta_{F_{ij}}| \geq \Delta \text{ or } |\phi_{ij}-\phi_{F_{ij}}| \geq \Delta \\
      \Omega_{\bar{z}_j^i} & \text{ otherwise } \\
\end{cases}
\end{align}
\noindent The error threshold $\Delta$ is determined based on the noise of the AOA profile which is shown to be Gaussian in~\cite{AURO2017} in theory with a computable variance. Here the error in the AOA profiles was found to be on the order of \ang{8.5} in our experimental study (see Figure~\ref{fig:aoa}). The weight $\mathcal{W}_{ij}$ is given by Equation \eqref{eq:w}. After properly adjusting the information matrix for each factor in Equation \eqref{eq:internal_pose_error}, the optimization result demonstrates smaller ground truth error. Figure~\ref{fig:csi_rejection} demonstrates hardware results showing improved ground truth error in optimized pose estimates using the information matrix from Equation~\eqref{eq:omega}.

\section{Results}
\vspace{-11.2 pt}
In this section, we describe our simulation and hardware experiments. For each experiment we compare the difference in performance with and without \emph{Active Rendezvous}. We employ a random exploration strategy as a benchmark though we note that our methods are fully compatible with general exploration methods as well. 
We utilize the Absolute Trajectory Error (ATE) metric
~\cite{sturm2012benchmark} 
and use a specialized version defined in~\cite{choudhary2017distributed}:
\setlength{\belowdisplayskip}{3pt}
\setlength{\abovedisplayskip}{2pt}
\begin{align}
    \label{eq:ATEtran}
    ATE_{\text{trans}} & := \frac{1}{\sum_{i=1}^k n_i}\sum_{i=1}^k \sum_{t=t_0}^{t_{n_i}} ||p_i(t) - \hat{p}_i(t) ||^2 \\
    \label{eq:ATErot}
    ATE_{\text{rot}} & :=  \frac{1}{\sum_{i=1}^k n_i} \sum_{i=1}^k \sum_{t=t_0}^{t_{n_i}} || R_i(t)^{\intercal}\hat{R}_i(t)-I_{3} ||_F^2
\end{align}
\begin{wrapfigure}[16]{r}{0.38\textwidth}    
\vspace{-23pt}
  \centering
    \includegraphics[width=0.38\textwidth]{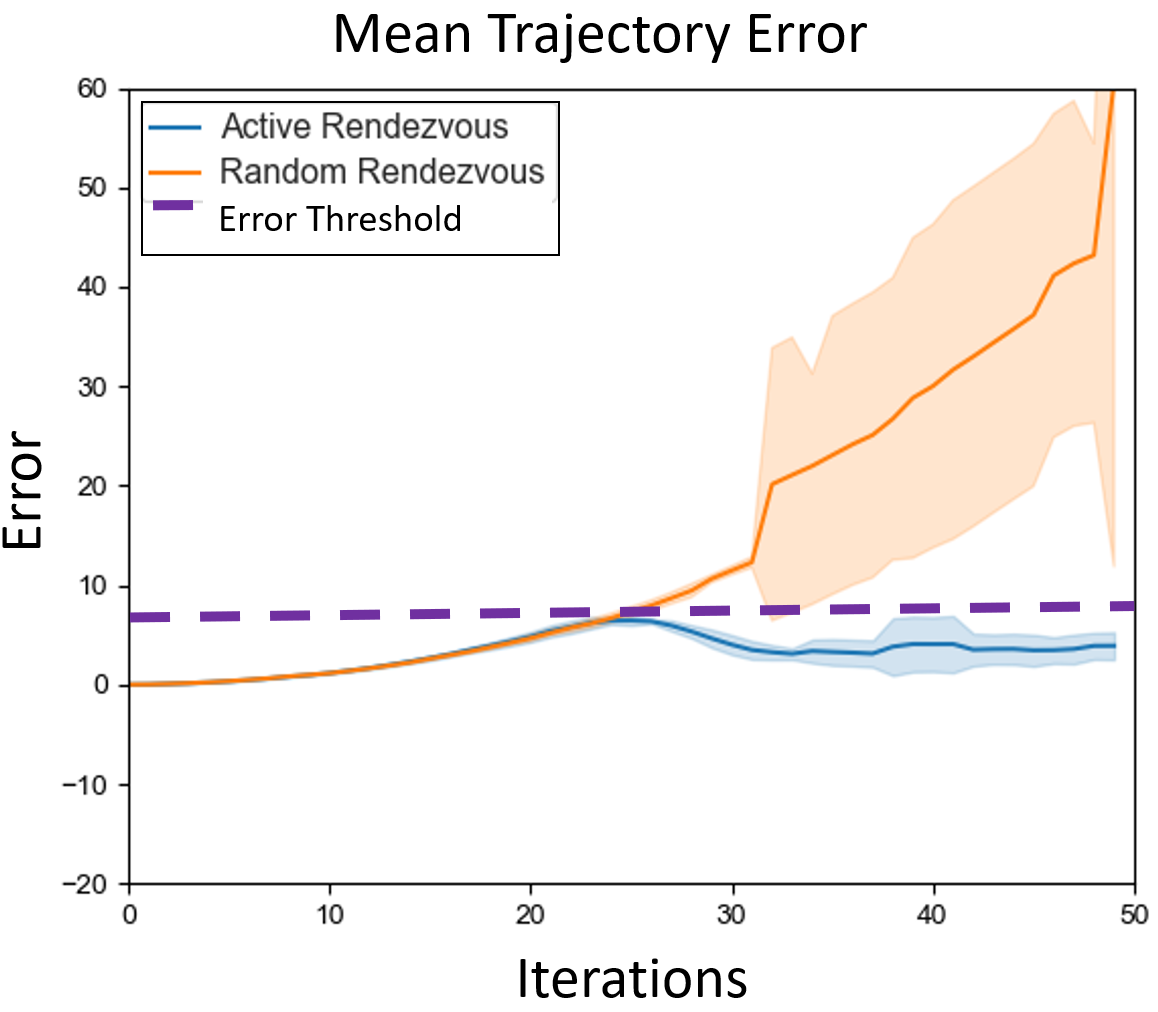}
  \vspace{-23pt}
  \caption{{\footnotesize{\emph{Results for 50 robot simulation with trajectory error bounded at desired threshold. Y-axis Error=$Err(x(t), \mathcal{E}_{\bar{z}}(t))$ and shaded region demonstrates standard deviation over 5 runs.}}}}
  \vspace{-25pt}
  \label{fig:simulation_results}
\end{wrapfigure}
\noindent where $k$ is the number of robots, $n_i$ is the number of poses in the trajectory of robot $i$, $p_i(t)$ and $R_i(t)$ are the estimated translation and rotation of robot $i$ at time $t$, and $\hat{p}_i(t)$ and $\hat{R}_i(t)$ are benchmark positions and rotations of robot $i$, typically according to ground truth from motion capture. In order to keep our methods general to different SLAM or PGO implementations we do not assume that observations are acquired with any particular sensor (i.e. camera, lidar, etc), but instead emulate them by using oracle positions (obtained from motion capture) with added Gaussian noise. Likewise, outlier measurements are emulated using Gaussian noise with large standard deviations compared to the measurement noise model. Although in practice outlying measurements can occur for a variety of reasons, such as erroneous loop closures or data aliasing, etc., we do not make assumptions about them or how they occur. We use identical PGO implementation across experiments with different rendezvous strategies (\emph{Random} vs. \emph{Active Rendezvous}); our results thus compare the difference in performance between using \emph{Active Rendezvous} and CSI-enabled outlier rejection and not using either. In particular our results show that by monitoring trajectory error as in Algorithm~\ref{alg:monitor} and enforcing \emph{Active Rendezvous} when it reaches a user-specified threshold, we can reduce the total trajectory error by more than 6X (Figure~\ref{fig:HW_active_trajectory}). This translates to an improvement in ground truth pose accuracy of 64\% (Figure~\ref{fig:gt_errors_hw}) as demonstrated in hardware experiments in a large environment (Testbed 2 is depicted in Figure~\ref{fig:hardware_setup}(c)). Using AOA data to assist in outlier rejection as presented in Section~\ref{sec:outliers} further reduces groundtruth pose error by 32\% (Figure~\ref{fig:csi_rejection}).\\

\vspace{-0.8cm}
\subsection{Simulation Experiments}
\label{sec:simulation}
\vspace{-0.25cm}
\subsubsection{Setup:}  

Our simulation consists of a framework developed using \emph{GTSAM} library, \emph{ROS} and \emph{Gazebo} for robot control and visualization. We first verify that the framework's backend optimizer, an implementation from authors of~\cite{choudhary2017distributed}, works well with the nearest neighbor selection in Algorithm ~\eqref{alg:monitor} for a large number of robots during \emph{Active Rendezvous}, as shown in Fig. \ref{fig:simulation_results}. The selection here uses trajectory error, robot ID data and a mapping of Euclidean distance to ESNR so as to approximate the behavior of communication quality in real wireless signals. Next, we test robot controls for obstacle detection and avoidance in Gazebo. 
The implementation is independent of any specific sensor-based SLAM approach as we instead added Gaussian noise to oracle observations.
\vspace{-0.5cm}
\subsubsection{Results:}
In order to examine the impact of using \emph{Active Rendezvous}, we first compare the trajectory error of our approach with a custom random exploration and rendezvous strategy in the simulation environments.To test the framework's scalability, we performed 5 trials with environment size of 10000 sq meters, each with 50 robots, that ran for 50 iterations, with estimated trajectory error threshold $\delta=10$. We add Gaussian noise of 0.2 meters in translation and \ang{5} in rotation for generating poses and measurements. Each robot is capable of moving 2 meters in one iteration and has a virtual sensor range of 2.2 meters to generate measurements. From Figure \ref{fig:simulation_results}, the aggregate results shows that our approach periodically enforces rendezvous over the experiment duration, providing better performance once the trajectory error first reaches the user-provided threshold. The average simulation time for each run was 5 minutes.
\vspace{-0.3cm}
\subsection{Hardware Experiments}
\begin{figure}[h]
    \centering
    \vspace{-0.95cm}
    \includegraphics[width=1.0\linewidth, height=4cm]{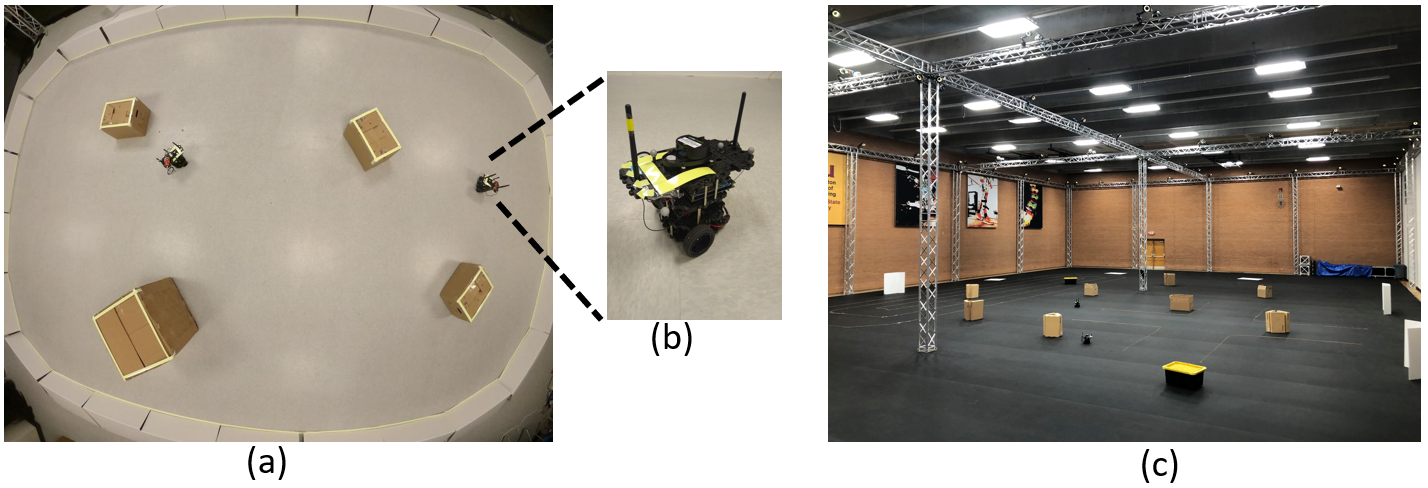} 
    \title{CSI based Outlier Rejection}
    \vspace{-20pt}
    \caption{\footnotesize{\emph{Hardware setups used for experimental evaluation: (a) Overhead view of 74 sq m Testbed 1, (b) Turtlebot3 robot platform (c) 3000 sq m Testbed 2}}}
    \label{fig:hardware_setup}
\end{figure}
\subsubsection{Setup:}

For hardware experiments, two \emph{Turtlebot3 Burger} robots were deployed in two testbeds of different sizes shown in Figure~\ref{fig:hardware_setup}, including a very large 3,000 square meter testbed (Figure~\ref{fig:hardware_setup} (c)). Each robot was equipped with  \emph{MinnowBoard Turbot Dual and Quadcore}  off-the-shelf SBCs (single board computers) running Ubuntu 16.04 LTS with kernel v4.15 and two $5dBi$ antennas spaced 22 cm apart, communicating over a 5 GHz channel. We attached an Intel 5300 Wi-Fi card to the Turbot which estimates wireless channels for each antenna via the 802.11n \emph{Channel State Information}(CSI) tool \cite{Halperin_csitool}. A Matlab framework calculated wireless signal profiles (Sec.~\ref{sec:primer}) once per call to Algorithm~\ref{alg:rendezvous} using ping packets collected over a quarter turn rotation (<1 second worth of data collection per profile) where the rotation angle of the robot was monitored using the native Turtlebot3 IMU sensor or \emph{OptiTrack} motion capture data. Algorithm~\ref{alg:rendezvous} then provided actuation commands to robots during \emph{Active Rendezvous}. Groundtruth pose data was collected using \emph{OpiTrack}. The \emph{Turbots} were set in monitor mode to broadcast fixed length ping packets at a rate of 200 packets/sec. 
The measurements $\mathcal{E}_{\bar{z}}^i(t)$ were generated by injecting noise from a zero-mean Gaussian distribution with standard deviation $\sigma_R = \ang{5}$ for rotation and $\sigma_t=$0.2 meters for the translations to groundtruth measurements similar to Sec.~\ref{sec:simulation}.

\vspace{-0.51cm}
\subsubsection{Results:}
\begin{wrapfigure}[20]{r}{0.51\textwidth}
\centering
   \vspace{-25pt}
    \includegraphics[width=0.5\textwidth]{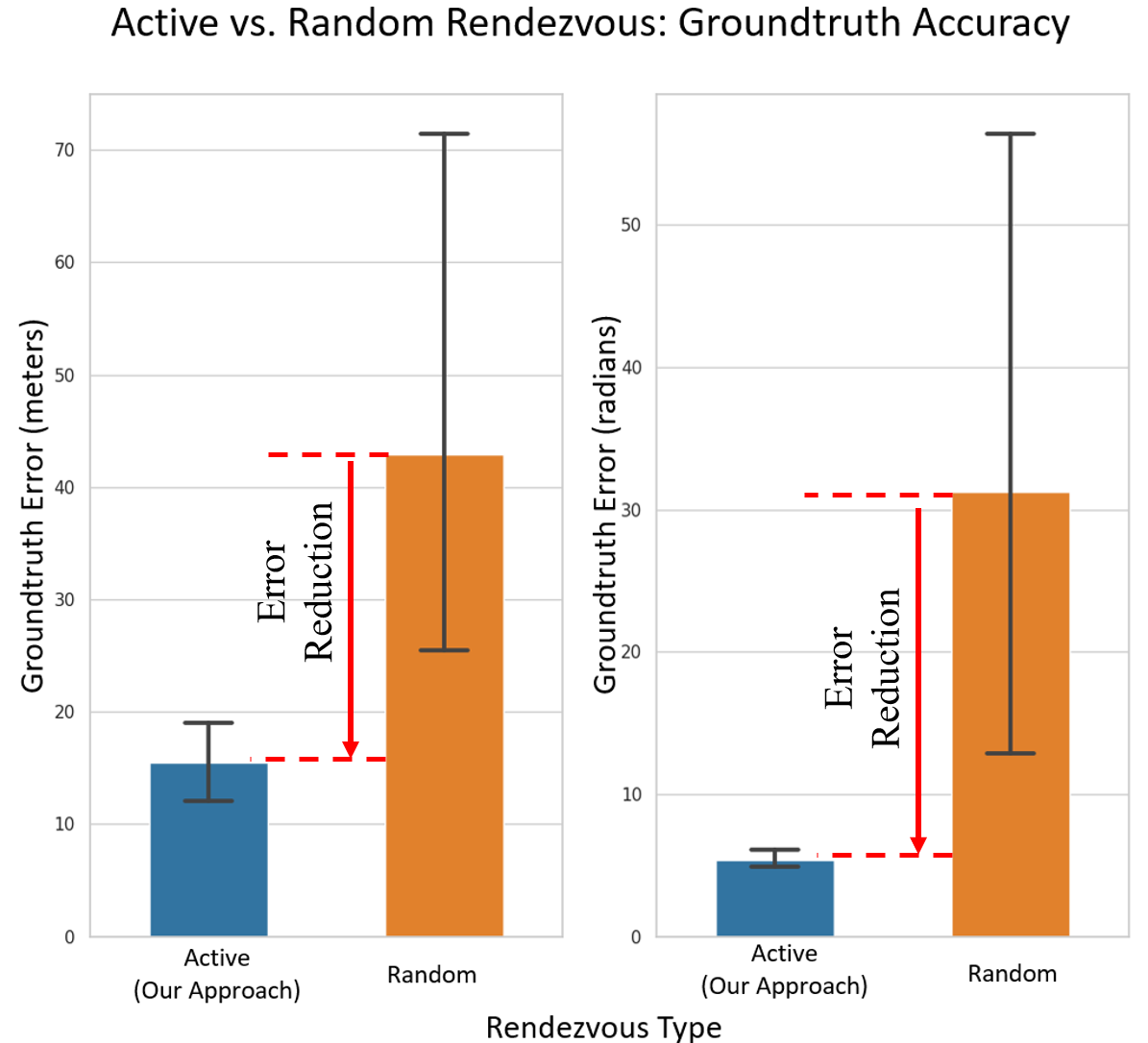}
  \vspace{-12pt}
  \caption{\footnotesize{\emph{Aggregate ground truth error (Equations~\eqref{eq:ATEtran}~\eqref{eq:ATErot}) reduction of $64\% $ on average over 3 hardware runs in large 3000 square meter testbed (Fig.~\ref{fig:hardware_setup}(c)). Average distance traversed is 100 meters for each robot. The low probability of random rendezvous leads to high variance in its groundtruth error. }}}
  \label{fig:gt_errors_hw}
\end{wrapfigure}
\begin{figure}[h]
    \centering
    \vspace{-15pt}
    \includegraphics[width=0.95\linewidth, height=6cm]{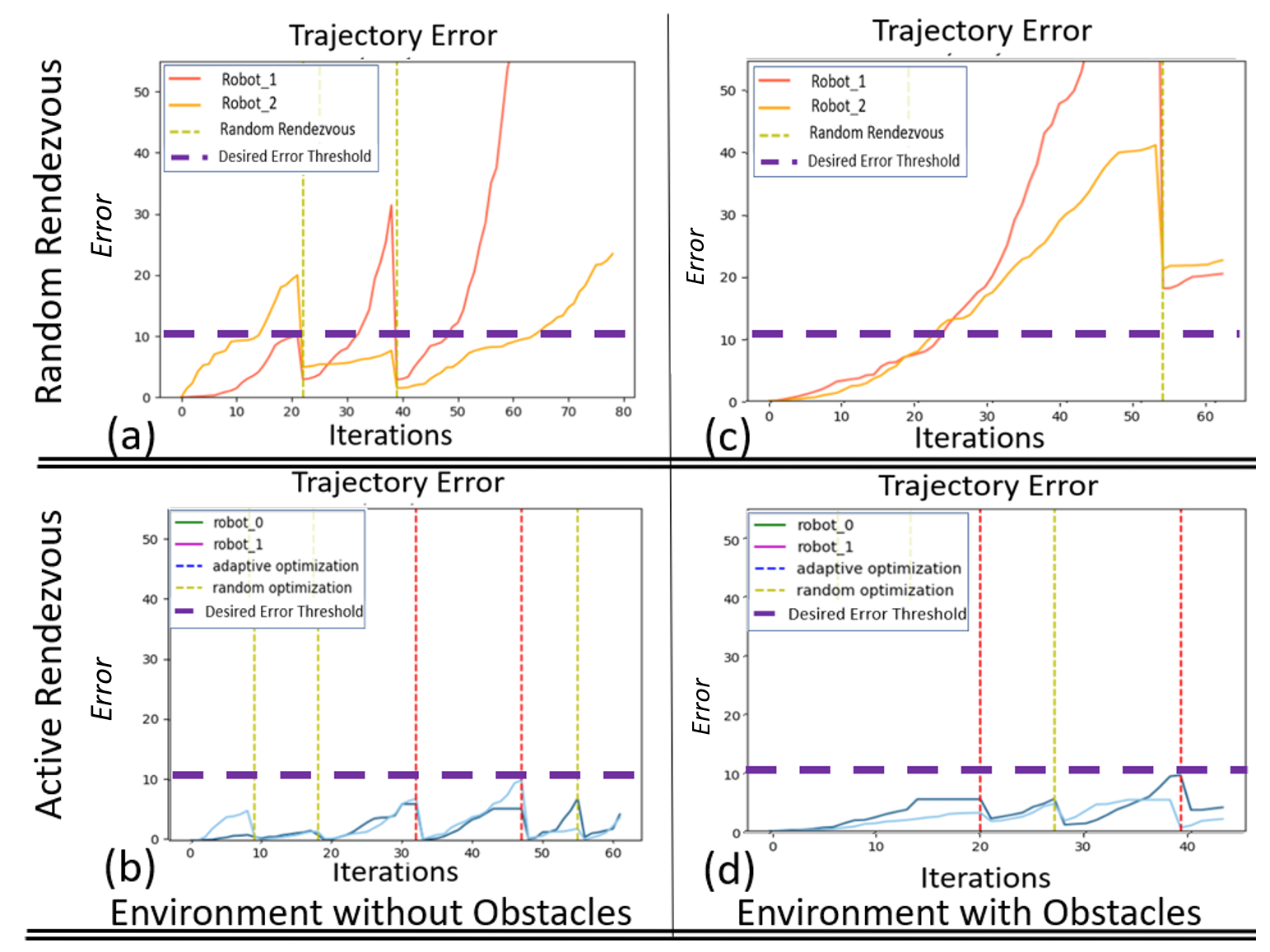}
    \vspace{-8pt}
    \caption{\footnotesize{\emph{Single trial of Random Rendezvous versus Active Rendezvous shows the growth of trajectory error over time due to noisy sensors (odometers in this case). Y-axis Error=$Err(x(t), \mathcal{E}_{\bar{z}}(t))$ from Equation~\eqref{eq:internal_pose_error}. Active Rendezvous reduces estimated trajectory error by 6X versus Random Rendezvous for different environments and platforms with heterogeneous sensor noise (Testbed 1, Fig~\ref{fig:hardware_setup}(a) is used here). Horizontal dashed lines indicate desired error threshold. Vertical lines indicate rendezvous.}}}
        \vspace{-20pt}
    \label{fig:HW_active_trajectory}
\end{figure}
\begin{figure}[h]
    \centering
    \vspace{-20pt}
    \includegraphics[width=1.0\linewidth, height=4cm]{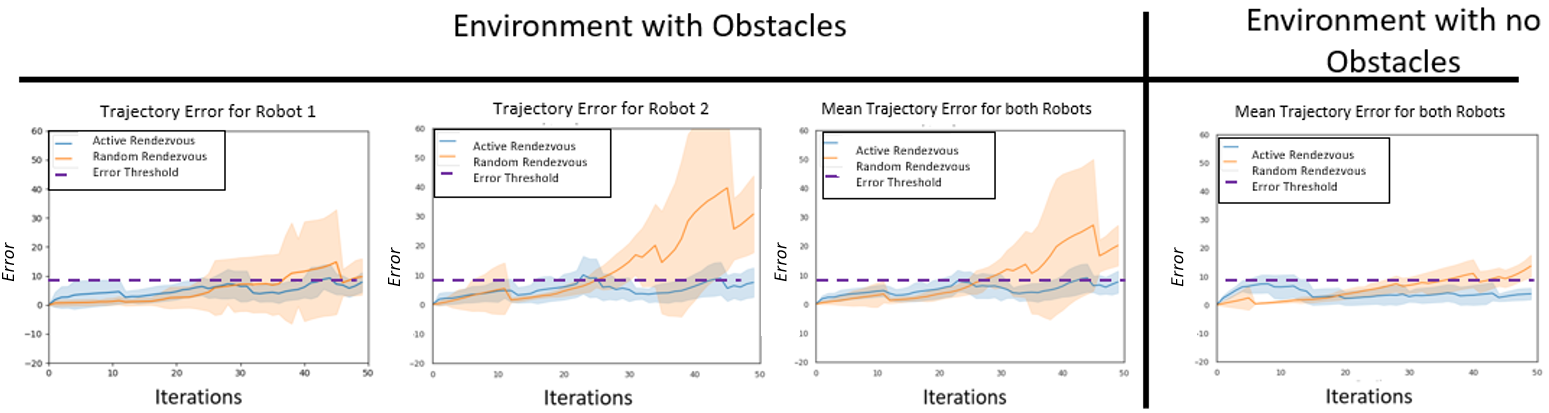}
    \vspace{-20pt}
    \caption{\footnotesize{\emph{Aggregate experimental results demonstrates bounded trajectory error for Active Rendezvous over 10 trials of hardware experiments in Testbed 1 from Fig~\ref{fig:hardware_setup}(a). Y-axis Error=$Err(x(t), \mathcal{E}_{\bar{z}}(t))$, shaded region indicates standard deviation. }}}
     \label{fig:Hardware_Aggregate}
\end{figure}
\begin{figure}[h]
\centering

    \subfloat[Error before and after rendezvous in well-lit conditions.]{\includegraphics[width=1.0\linewidth, height=3.5cm]{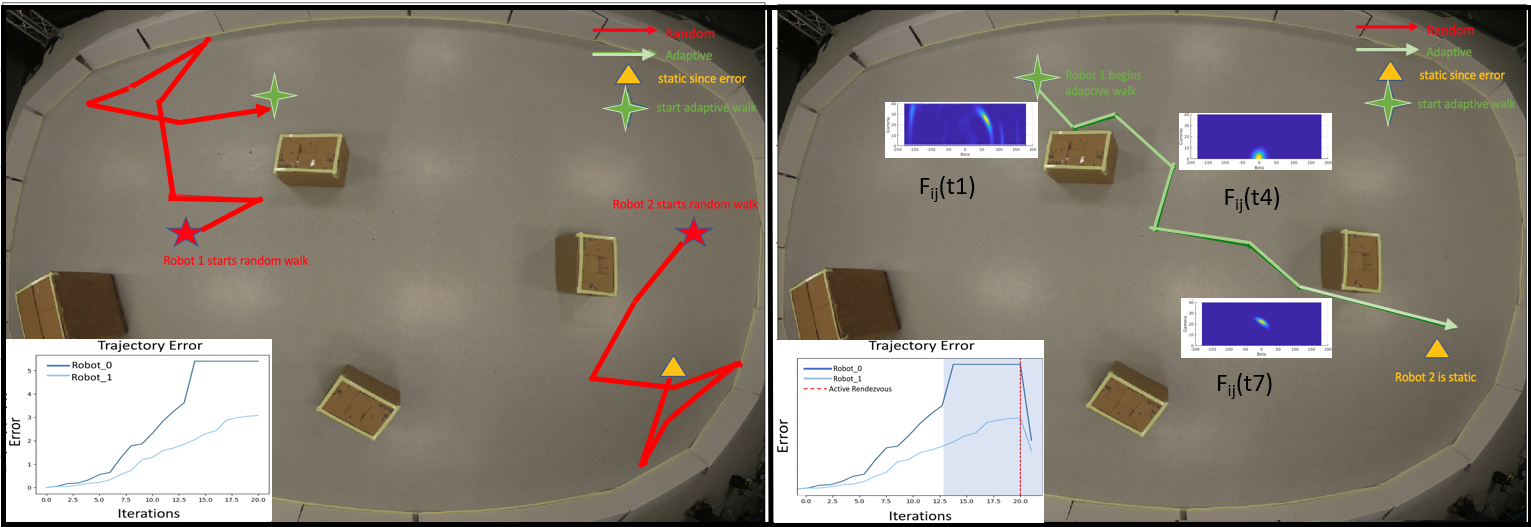}}\\
    \subfloat[Error before and after rendezvous in low-lighting conditions.]{\includegraphics[width=1.0\linewidth, height=4cm]{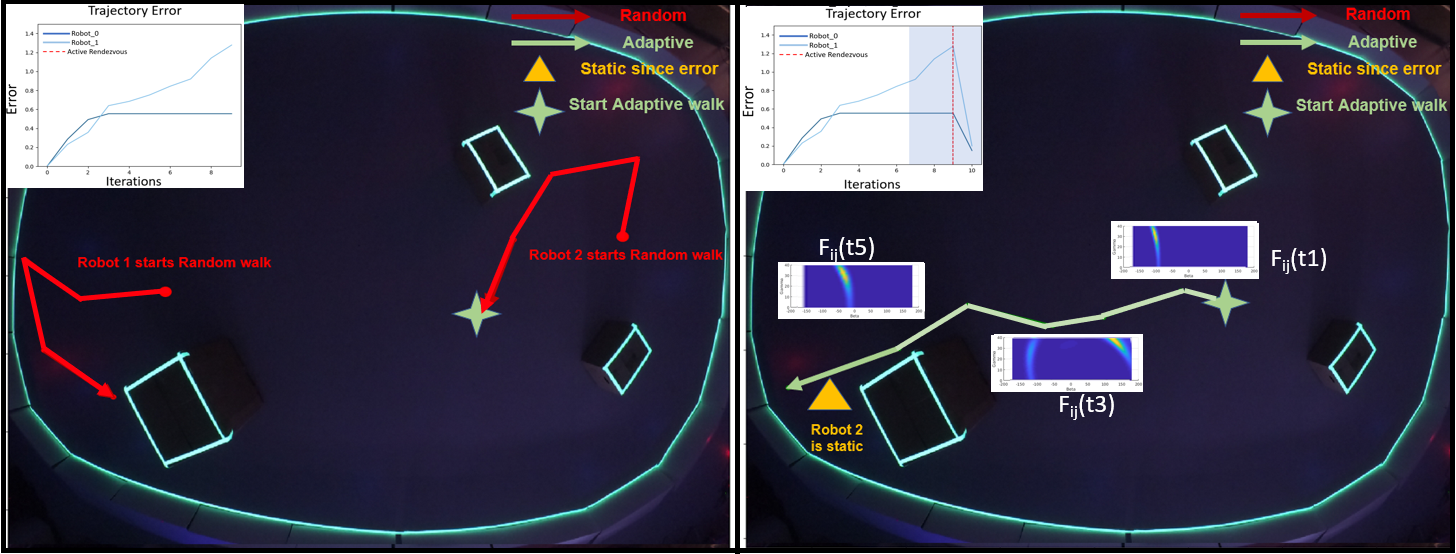}}
    \vspace{-10 pt}
    \caption{\footnotesize{\emph{Active Rendezvous in a visually degraded environment where vision based rendezvous would fail. Y-axis Error=$Err(x(t), \mathcal{E}_{\bar{z}}(t))$. Testbed boundary is shown fluoresced (not visible to robots). Shaded regions of plots show adaptive navigation and Active Rendezvous during one instance of optimization.}}}
    \label{fig:HW_active_trajectory_dark_room}
\end{figure}

We compare our approach to only using random rendezvous with three sets of real-world experiments 
for validation in hardware settings and to demonstrate our \emph{Active rendezvous} approach works in the presence of obstacles. The initial two experiments were carried out in testbed shown in Figure~\ref{fig:hardware_setup}(a) for 50 iterations with each robot moving 1 meters during an iteration and having an observation distance of 1.1 meters. We first display the estimated trajectory error $Err_i(x_i(t),\mathcal{E}_{\bar{z}}^i(t))$ and rendezvous history for a single trial of hardware experiments in an environment with and without obstacles. It can be seen from Fig. \ref{fig:HW_active_trajectory} that early in the trial, the random rendezvous based approach performs similarly to our approach due to several random rendezvous. However, after a long period without a rendezvous, the former begins to diverge. In contrast, our method begins to enforce \emph{Active Rendezvous} as soon as the trajectory error increases. Additionally, examining Fig. \ref{fig:HW_active_trajectory} (c),(d), we note that the presence of obstacles affects the number of rendezvous. 
The next set of experiments were performed in the environment shown in Figure ~\ref{fig:hardware_setup}(a) without obstacles. We performed 6 trials that each ran for 50 iterations. We then introduced 4 obstacles and performed the same number of trials and iterations per trial in this environment. Aggregate results, as well as individual robot errors, for both experimental conditions are shown in Fig.~\ref{fig:Hardware_Aggregate}. We see the same trends as in simulation: early on in the experiments, both methods show similar performance, but over time the random-walk based method degrades while our approach remains accurate. Finally, we performed identical hardware experiments, both with and without obstacles, in extreme low light conditions (Fig. \ref{fig:HW_active_trajectory_dark_room}). These experiments were designed to showcase that our method of rendezvous is completely independent of visual conditions, in a dark, featureless environment which would severely inhibit vision-based place recognition algorithms for coordinating rendezvous. The last set of experiments were carried out in a larger testbed - Figure ~\ref{fig:hardware_setup}(c), to evaluate the impact of \emph{Active Rendezvous} on groundtruth errors calculated per~\cite{Zhang18iros}. Each robot moved an average distance of 100 meters during every run, using an observation distance of 2.2 meters to generate measurements. 3 such runs were conducted, each lasting 20 minutes on average. The aggregate results in Figure~\ref{fig:gt_errors_hw} shows that enforcing rendezvous due to bounds on trajectory error, our approach significantly increases accuracy while remaining agnostic to the exploration method of choice and the environment.

\begin{figure}[h]
    \centering
    \vspace{-10pt}
    \includegraphics[width=1.0\linewidth, height=5.5cm]{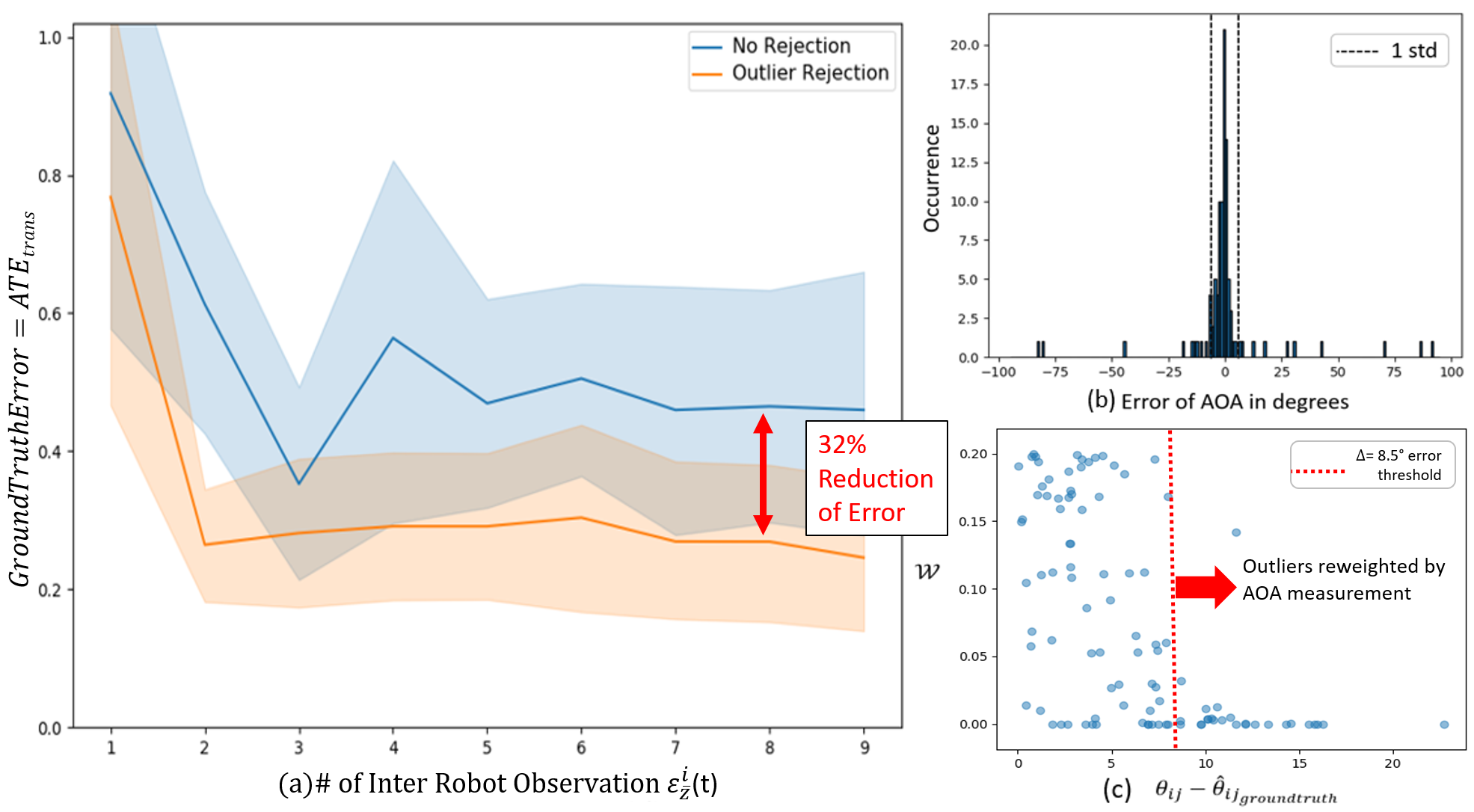} 
    \title{CSI based Outlier Rejection}
    \vspace{-20pt}
    \caption{\footnotesize{\emph{(a)Aggregate results over 5 independent experiments carried out in 74 sq meter testbed (Fig.~\ref{fig:hardware_setup}(a)) show 32\% reduction in groundtruth error by using AOA for outlier rejection. (b)Error distribution of AOA information in $F_{ij}$ shown to be approximately Gaussian for actual profiles obtained using native IMU on \emph{Turtlebot3} platform. (c) Weights as computed from Eqn.~\eqref{eq:w} and used for outlier rejection in Eqn.~\eqref{eq:omega} are shown to correctly reduce the weighting for outlier observations differing from ground truth for any $\Delta\geq\ang{8.5}$ in our implementation.}  }}
    \label{fig:csi_rejection}
\end{figure}
\subsection{Experimental Results for Outlier Rejection}
\vspace{-0.3cm}
In this section we present hardware results for using CSI to detect and weight outlier relative pose measurements prior to performing optimization. The  measurements are injected with random translation and rotation error between (2$\sigma$ 3$\sigma$) of the original noise distribution to generate outliers.As explained in Sec.~\ref{sec:outliers}, we collect 9 measurements between robots for each of the 5 trials of \emph{Active Rendezvous}. Figure~\ref{fig:csi_rejection} (a),(b) and (c) show the average groundtruth error from Equation ~\eqref{eq:ATEtran} in estimated trajectory  before and after outlier rejection, error distribution of 100 AOA  samples and the joint-distribution of AOA and measurements respectively. The $\mathcal{W}$ in Figure \ref{fig:csi_rejection} (c) is computed from Equation~\eqref{eq:w}. This shows that the AOA measurements reliably detect outliers for $\Delta\geq\ang{8.5}$ as a rejection boundary. Figure~\ref{fig:csi_rejection} (a), shows that using AOA to reject observations leads to a significant decrease in average groundtruth error. This reduction highlights the benefit of having an independent modality to validate measurements.

\vspace{-0.45cm}
\section{Conclusion and Future Work}
\vspace{-0.45cm}
We have presented a distributed method for integrating directional information from wireless communications into pose graph optimization that enables robots to rendezvous with one another in a way that is \emph{independent of the environment and is capable of rejecting outlying relative pose observations}.  We demonstrate the utility of this method both in simulation and in hardware experiments. We attain an improvement in ground truth pose accuracy of 64\% with \emph{Active Rendezvous} compared to random rendezvous using similar exploration strategies. Additionally, using CSI information in outlier rejection improves ground truth accuracy by 32\%. Future work could involve integrating this information directly in the pose graph optimization.

{\scriptsize \noindent\textbf{Acknowledgements:} The authors gratefully acknowledge support by the NSF CAREER award number 1845225, MIT Lincoln Labs Line Grant, and the Fulton Undergraduate Research Initiative.}\\
{\scriptsize \noindent \textbf{Disclaimer:} DISTRIBUTION STATEMENT A. Approved for public release. Distribution is unlimited. This material is based upon work supported by the Under Secretary of Defense for Research and Engineering under Air Force Contract No. FA8702-15-D-0001. Any opinions, findings, conclusions or recommendations expressed in this material are those of the author(s) and do not necessarily reflect the views of the Under Secretary of Defense for Research and Engineering.}

\renewcommand*{\bibfont}{\footnotesize}

\bibliographystyle{spbasic}
\renewcommand{\bibname}{References}
\vspace{-0.4cm}

{
\bibliography{references}}

\end{document}